\pdfoutput=1
\documentclass[runningheads]{llncs}

\usepackage[utf8]{inputenc} 
\usepackage[colorlinks]{hyperref}       
\usepackage{url}            
\usepackage{booktabs}       
\usepackage{amsfonts}       
\usepackage{nicefrac}       

\usepackage{graphicx}
\usepackage{pgf}
\usepackage{color}
\usepackage{subcaption}
\usepackage{tabularx}
\usepackage{colortbl}
\usepackage{multirow}
\usepackage{enumitem}
\usepackage{etoolbox}
\usepackage[capitalize]{cleveref}

\usepackage[absolute]{textpos}

\newcommand{\legendentry}[2]{\textcolor{#2}{\rule[1.5pt]{10pt}{.75pt}} \enskip #1}

\definecolor{gray92}{gray}{0.92}
\definecolor{gray96}{gray}{0.96}
\definecolor{pltclr1}{rgb}{0.121569,0.466667,0.705882}
\definecolor{pltclr2}{rgb}{1.000000,0.498039,0.054902}
\definecolor{pltclr3}{rgb}{0.172549,0.627451,0.172549}
\definecolor{pltclr4}{rgb}{0.839216,0.152941,0.156863}
\definecolor{pltclr5}{rgb}{0.580392,0.403922,0.741176}
\definecolor{pltclr6}{rgb}{0.549020,0.337255,0.294118}
\definecolor{correct}{rgb}{0.2, 0.8, 0.0}
\definecolor{wrong}{rgb}{0.7, 0.13, 0.13}

\newcolumntype{s}{>{\columncolor{gray92}}c}
\newcolumntype{m}{>{\columncolor{gray96}}c}
\newcolumntype{C}{>{\centering\arraybackslash}X}

\newtoggle{blind}
\newtoggle{tweethist}
\togglefalse{blind}
\toggletrue{tweethist}

\begin{document}
    
\title{Finding Relevant Flood Images on Twitter\\using Content-based Filters}

\titlerunning{Finding Relevant Flood Images on Twitter using Content-based Filters}

\iftoggle{blind}{
    
\author{Blind submission}
\authorrunning{Anonymous MAES submission}
\institute{}

}{

\author{Bj{\"o}rn Barz\inst{1} \and
    Kai Schr{\"o}ter\inst{2} \and
    Ann-Christin Kra\inst{2} \and
    Joachim Denzler\inst{1}}
\authorrunning{B. Barz et al.}
%
\institute{Computer Vision Group, Friedrich Schiller University Jena, Jena, Germany\\
    \email{\{bjoern.barz,joachim.denzler\}@uni-jena.de}\and
    Section of Hydrology, Deutsches GeoForschungsZentrum, Potsdam, Germany\\
    \email{\{kai.schroeter,annkra\}@gfz-potsdam.de}}

}

\maketitle

\begin{textblock*}{140mm}(37mm,20mm)
    \textblockcolour{gray92}
    \vspace{2mm}
    \tiny
    \centering
    B. Barz, K. Schr{\"o}ter, A.-C. Kra, J. Denzler.\\
    Finding Relevant Flood Images on Twitter using Content-based Filters.\\
    ICPR 2020 Workshop on Machine Learning Advances Environmental Science, 2020.\\
    \copyright\ Copyright by Springer. The final publication is available at
    \href{https://link.springer.com/}{link.springer.com}.\\
    \vspace{2mm}
\end{textblock*}

\begin{abstract}
  The analysis of natural disasters such as floods in a timely manner often suffers from limited data due to coarsely distributed sensors or sensor failures.
  At the same time, a plethora of information is buried in an abundance of images of the event posted on social media platforms such as Twitter.
  These images could be used to document and rapidly assess the situation and derive proxy-data not available from sensors, e.g., the degree of water pollution.
  However, not all images posted online are suitable or informative enough for this purpose.
  
  Therefore, we propose an automatic filtering approach using machine learning techniques for finding Twitter images that are relevant for one of the following information objectives: assessing the flooded area, the inundation depth, and the degree of water pollution.
  Instead of relying on textual information present in the tweet, the filter analyzes the image contents directly.
  We evaluate the performance of two different approaches and various features on a case-study of two major flooding events.
  Our image-based filter is able to enhance the quality of the results substantially compared with a keyword-based filter, improving the mean average precision from 23\% to 53\% on average.
  
\end{abstract}

\section{Introduction}

Floods cause severe damages in urban areas every year.
Up-to-date information about the flood is crucial for reacting quickly and appropriately by providing assistance and coordinating disaster recovery \cite{comfort2004coordination}.
However, traditional sensors such as water level gauges often fail to provide the required information \cite{thieken2016flood}, either because of failures, a too coarse spatio-temporal resolution, or even the absence of sensors, e.g., in the case of surface water flooding, which can occur far off the next stream.

For a rapid flood impact analysis and documentation, more versatile sources of information are hence desirable for augmenting the data obtained from sensors.
One such source of information, which is available in abundance nowadays, consists in images of the flood posted by citizens on social media platforms.
Thanks to the widespread adoption of smartphones, this information is usually available rapidly during the flood event and covers populated areas with a better resolution than traditional sensors.
Though images do not provide accurate measurements, as opposed to sensors, they are in many cases still useful to estimate important information.
For example, the boundaries of the flooded area can easily be identified in an image.
The grade of water pollution can also be assessed visually for certain types of pollution such as, e.g., oil spills.
It can even be possible to estimate the inundation depth if the image shows objects of known height in the flooded area, such as traffic signs, cars, or people.

Fohringer et al.~\cite{fohringer2015social} made use of this information by manually inspecting the images of all tweets that contained a flood-related keyword and were posted in the area and timeframe of interest.
Beyond simple keyword matching, sophisticated machine learning techniques have recently been applied to filter event-related tweets based on their text \cite{kruspe2019detecting}.
These text-based approaches have two major drawbacks when it comes to finding images of the event:
First and foremost, a text-based filter captures far too many images for manual inspection in the context of rapid flood impact assessment.
Secondly, users may not always mention the flood directly in the text of the tweet, since the topic is already recognizable from the image.
A text-based filter would miss these images and, therefore, ignore potentially useful information.

To overcome these issues, Barz et al.~\cite{barz2019enhancing} recently proposed a filtering technique based solely on the content of the images using an interactive image retrieval approach.
In their framework, the user initiates the process by providing an example image illustrating what they are looking for.
The system then retrieves a first set of similar images in which the user flags a handful of images as relevant or irrelevant to subsequently refine the search results over several feedback rounds.
Such an interactive approach is suitable in face of an open set of possible search objectives, since the system adapts to the user's needs from scratch during each session.
While this seems useful for a detailed post factum analysis of the event, the interactive feedback process is too time-consuming for rapid flood impact analysis during the flood.
Moreover, the set of important search objectives is usually more limited during this phase of analysis and focuses on a few key metrics such as the spread and depth of the flood.
In such a scenario, it is redundant to refine the system from scratch several times.
Instead, a classifier trained in advance on an annotated dataset can be used to filter the social media images of the event quickly without user interaction.

Therefore, this work focuses on developing a pre-trained non-interactive filter for relevant flood images.
We use the annotated European Flood 2013 dataset from Barz et al.~\cite{barz2019enhancing} as training data and demonstrate that the filters learned on those images collected from Wikimedia Commons also perform well in practice on Twitter images.
To this end, we evaluate their performance on images we collected from Twitter regarding two real flooding events, which occurred in 2017 and 2018 in Germany.
We find that such pre-trained filters clearly outperform a purely keyword-based approach, which ignores the image content.

\begin{figure}[t]
    \includegraphics[width=\linewidth]{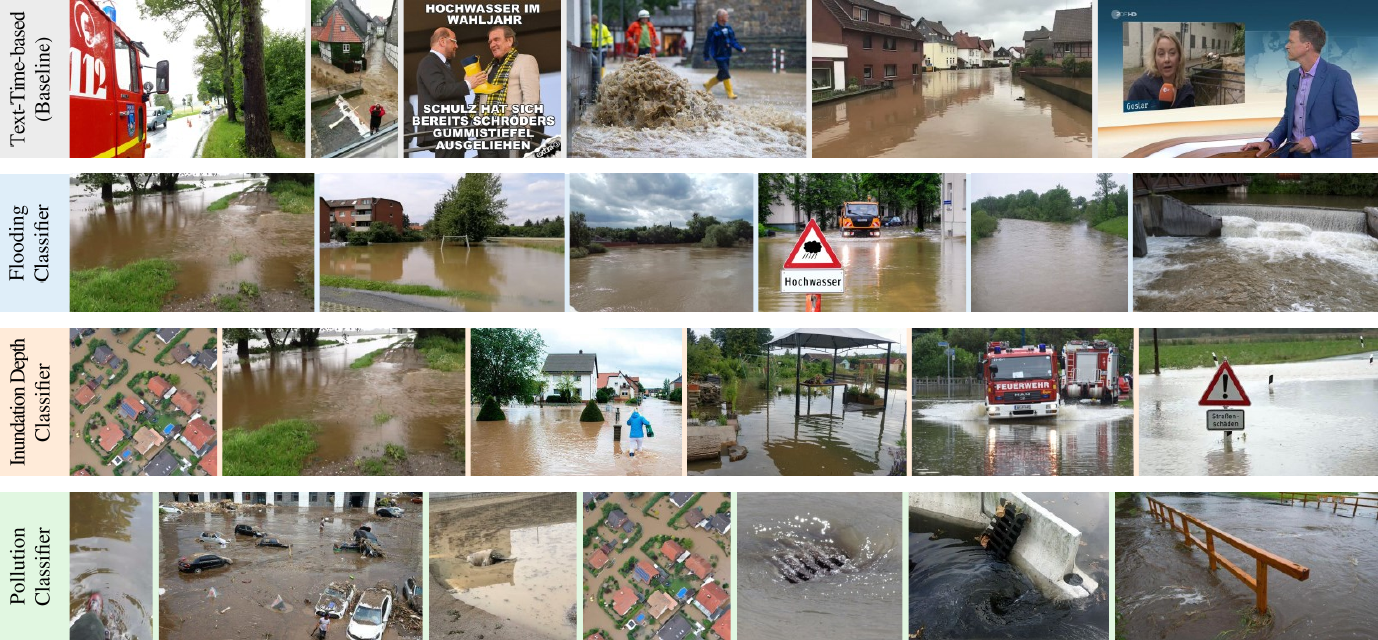}
    \caption{Comparison of na\"ive flood image retrieval based on textual keywords and the time of the tweet (top row) and three classification-based filters optimized for different information objectives (last three rows). Each row shows six top-scoring images from the Harz17 dataset.}
    \label{fig:examples}
\end{figure}

\Cref{fig:examples} shows an example illustrating the value of task-aware image-based filtering.
The results in the top row have been retrieved based on textual keywords and their temporal proximity to the peak of the event.
Besides flood images, they also contain irrelevant images without flooding, memes, and still images from news shows.
The following three rows, in contrast, show the top-scoring results obtained from three different filters:
The results of the \textit{flooding} filter mainly contain images depicting the boundary between flooded and non-flooded areas.
The \textit{inundation depth} filter focuses on visual clues that are helpful for determining the approximate depth of the flood such as traffic signs and people standing in the water.
Finally, the \textit{pollution} filter searches for images of heavily polluted water.

We describe these tasks and the datasets used in this study in more detail in the following \cref{sec:data} and then explain the different filtering approaches developed and tested in this work in \cref{sec:methods}.
Experimental results on the European Flood 2013 dataset and our two novel twitter datasets are shown in \cref{sec:experiments} and \cref{sec:conclusions} concludes this work.

Our two Twitter datasets and the best filter models are publicly available at \url{https://github.com/cvjena/twitter-flood-dataset}.

\section{Datasets and Search Objectives}
\label{sec:data}

We use an existing annotated dataset of flood images for training our filters and evaluate them on two novel flood datasets collected from Twitter.

\subsection{The European Flood 2013 Dataset}
\label{subsec:eu-flood-dataset}

The European Flood 2013 dataset \cite{barz2019enhancing} is a multi-label dataset comprising 3,710 images collected from the Wikimedia Commons category ``Central Europe floods, May/June 2013''.
Each image has been annotated by experts from hydrology regarding its relevance with respect to the following three search objectives:

\begin{description}[leftmargin=.8cm,topsep=0pt]
    \item[\textit{Flooding}] Does the image help to determine whether a certain area is flooded or not?
    An image considered as relevant would show the boundary between flooded and dry areas.
    Images that do not show any inundation at all are considered not relevant.
    
    \item[\textit{Inundation depth}] Is it possible to derive an estimate of the inundation depth from the image due to visual clues such as, for example, traffic signs or other structures with known height?
    If there is no flooding at all, the image is considered as not relevant for inundation depth.
    
    \item[\textit{Water pollution}] Does the image show any pollution substances?
    The focus is on heavy contamination by chemical substances such as oil, for example.
\end{description}

\noindent
For each of these objectives, between 100 and 250 images have additionally been selected as ``ideal queries'', which are considered to represent the information objective particularly well.
We will use these queries later to compare our pre-trained filters with a retrieval-based method (see \cref{subsec:retrieval}).

The dataset is typically augmented with 97,085 distractor images from the Flickr100k dataset \cite{Philbin07flickr100k} (excluding images with the tags ``water'' and ``river''), which are not considered as relevant for any of the aforementioned tasks.
We use the combined dataset of almost 100,800 images for our experiments and split it randomly into 75\% for training and 25\% for testing.

\subsection{Real-world Twitter Data}
\label{subsec:twitter-datasets}

To evaluate the performance of the methods investigated in this work in a realistic scenario, we collected images posted on Twitter during two major flood events in Germany:
The flood of July 2017 in the \textit{Harz} region caused severe damages to buildings, public infrastructure, and dikes in many cities in the center of Germany.
In January 2018, a flood of the river \textit{Rhine} in western Germany affected one of the largest German cities, Cologne, so that we can expect a high number of tweets relating to this flood.
We denote these two events as \textit{Harz17} and \textit{Rhine18} in the following.

We extracted potentially flood-related tweets during these two months from a database that we constructed using the Twitter Streaming API over the course of several years.
While our method is, in principle, purely image-based and does not rely on textual cues, the limitations of the API enforced us to pre-filter tweets based on the appearance of flood-related keywords\footnote{The German keywords were: Hochwasser, Flut, Überschwemmung, Überschwemmungen,  überschwemmt, überflutet, Sturmflut, Pegel.}.
However, the results still contain numerous unrelated tweets, as can be seen in the top row of \cref{fig:examples}.

\iftoggle{tweethist}{
\begin{figure}[t]
    \begin{subfigure}{.48\linewidth}%
        \caption{Harz17}%
        \label{subfig:histogram-harz17}%
        \includegraphics[width=\linewidth]{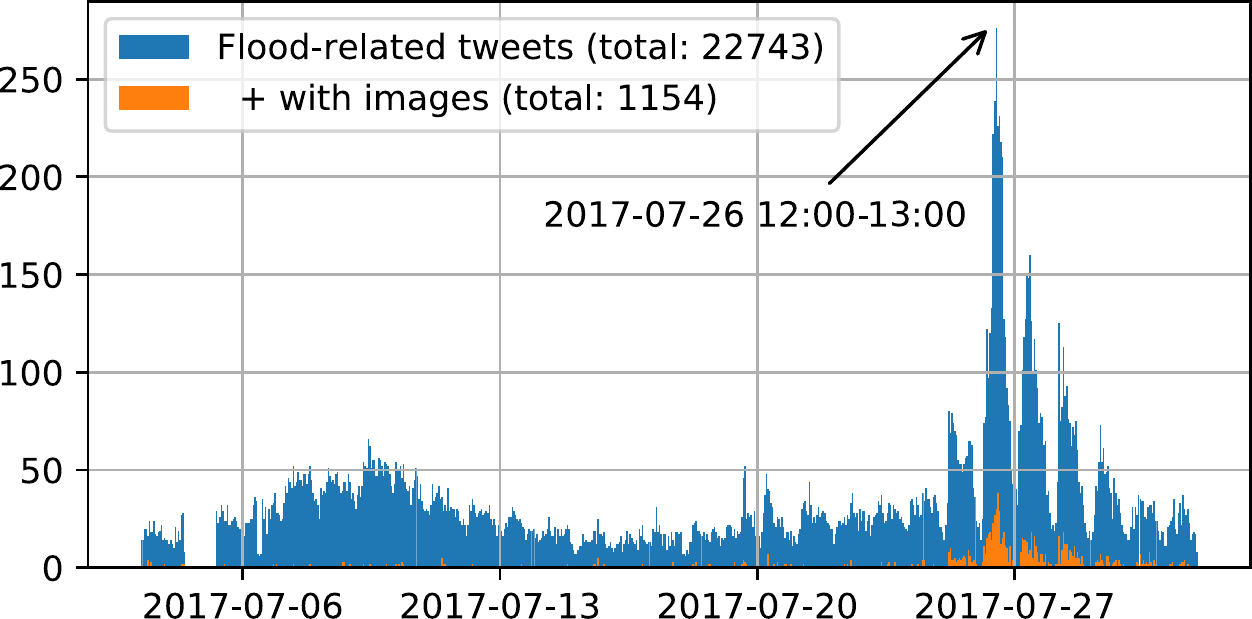}%
    \end{subfigure}%
    \hfill%
    \begin{subfigure}{.48\linewidth}%
        \caption{Rhine18}%
        \label{subfig:histogram-rhine18}%
        \includegraphics[width=\linewidth]{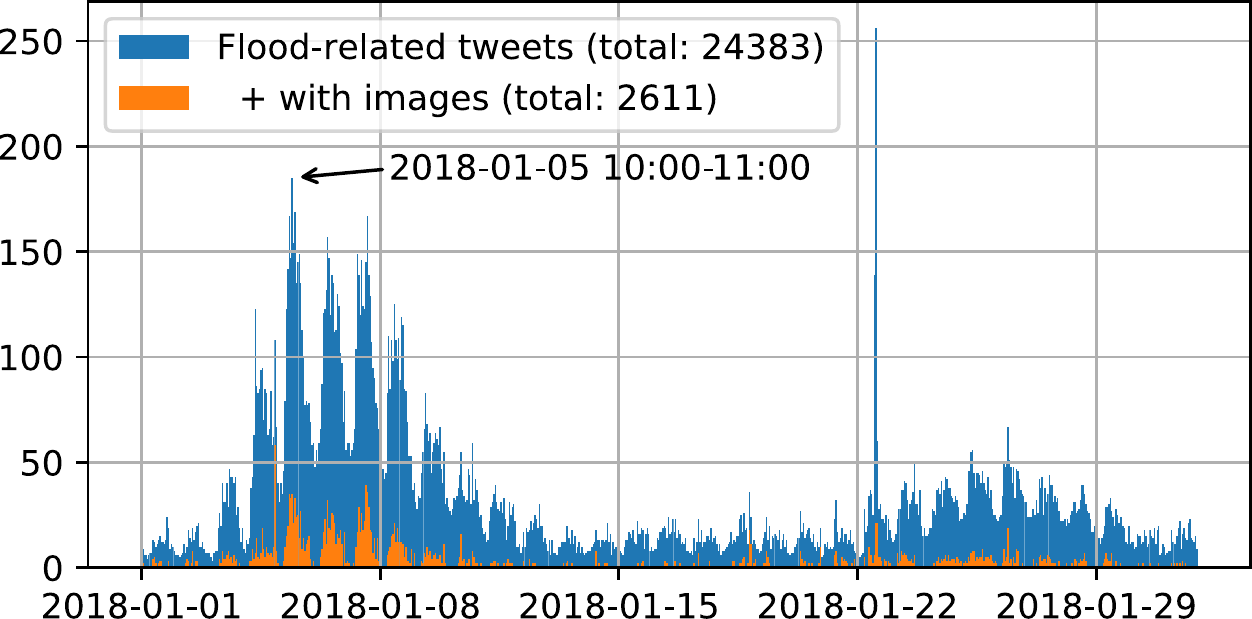}%
    \end{subfigure}%
    \caption{Tweets per hour matching a flood-related keyword during the two months used for our Twitter flood datasets.}
    \label{fig:twitter-histogram}
\end{figure}

\Cref{fig:twitter-histogram} depicts the number of tweets per hour containing a flood-related keyword during the two months under observation.
The increase in flood-related Twitter activity during late July 2017 and early January 2018 is clearly visible.
}{}

At the time of writing, 3,314 out of the 3,765 images posted on Twitter during these months were still accessible.
After a near-duplicate removal step using feature similarity and manual inspection of suspect duplicates, 704 images remain for the Harz17 dataset and 1,848 for the Rhine18 dataset.

We asked two experts from hydrology to annotate these datasets according to the same criteria as the European Flood 2013 dataset.
Due to the high number of images and limited resources, each image was annotated by a single expert.

\section{Methods}
\label{sec:methods}

We compare three methods for filtering relevant tweets: an objective-agnostic baseline relying on textual keywords and the date of the tweet, a retrieval approach, and a classification-based method.

\paragraph{Text-Time-based Baseline:}
\label{subsec:baseline}

As a na\"ive baseline ignoring the image contents, we rank all tweets containing a flood-related keyword by the proximity of their time of posting to the hour of maximum tweet frequency during the flood event.

\paragraph{Filtering by Retrieval:}
\label{subsec:retrieval}

In a general content-based image retrieval (CBIR) scenario, the user initially provides a set of query images represented by feature vectors $Q = \{q_1, \dots, q_m\} \subset \mathbb{R}^D$ and the system then ranks the images in the database by decreasing similarity of their feature vectors $X = \{x_1, \dots, x_n\} \subset \mathbb{R}^D$ to the queries.
Since we are not focusing on interactivity in this work but on a fixed set of search objectives, we relieve the user from the burden of specifying query images and fix $Q$ to the set of ``ideal queries'' for the respective task from the European Flood 2013 dataset.

For computing the similarity between an image feature vector $x \in X$ from the database and all queries, we use kernel density estimation (KDE), inspired by Deselaers et al.~\cite{deselaers2008learning}:%
\begin{equation}%
    \mathrm{sim}(x, Q) = \frac{1}{|Q|} \sum_{q \in Q} \exp\left(-\gamma \cdot \|x-q\|^2\right) \,.
\end{equation}%
The hyper-parameter $\gamma \in \mathbb{R}^+$ is tuned on the training set of the European Flood 2013 dataset.
The exact value can be found in \cref{tbl:hyperparams}.

To compute feature vectors describing the images, we employ convolutional neural networks (CNNs), which learned important image features such as, e.g., the appearance of certain shapes or textures, from large amounts of data.
First, we use features extracted from the last convolutional layer of a VGG16 architecture \cite{simonyan2014vgg} and a ResNet-50 architecture \cite{he2016resnet}, both pre-trained on 1.2 million images from ImageNet \cite{russakovsky2015ilsvrc}.
These regional features are averaged and finally $L^2$-normalized, resulting in a 512-dimensional feature space for VGG16 and 2048 features for ResNet-50.
The images are initially resized and cropped to $768 \times 512$ or $512 \times 768$ pixels, depending on their orientation.
This resolution corresponds to the median aspect ratio (3:2) in the Twitter datasets.

Secondly, we examine the use of features optimized for instance retrieval:
The Deep R-MAC architecture \cite{gordo2016end} is based on ResNet-101 and aggregates 2048-dimensional image features across several regions of interest and image resolutions, decorrelates them using PCA, and finally applies $L^2$-normalization.
It has been pre-trained on a landmark dataset using a metric learning objective that forces similar images to be closer than dissimilar images by a pre-defined margin.
This representation has been identified by Barz et al.~\cite{barz2019enhancing} to be more powerful than VGG16 features for interactive flood image retrieval, but they did not compare it with other ResNets trained on different datasets (e.g., ImageNet).

\paragraph{Filtering by Classification:}
\label{subsec:classification}

For each of the search objectives defined in \cref{subsec:eu-flood-dataset}, we train a binary linear support vector machine (SVM) \cite{cortes1995support} using all images from the European Flood 2013 dataset annotated as relevant for that task as positive example and all remaining images as well as the images from the Flickr100k dataset as negative examples.
We examine the same image features used for the retrieval technique (see above) and optimize the regularization hyper-parameter $C \in \mathbb{R}^+$ of the SVMs using 5-fold cross-validation on the training set.
The resulting value for all network architectures can be found in \cref{tbl:hyperparams}.

\begin{table}[b]
    \caption{Values of the hyper-parameters $\gamma$ for retrieval and $C$ for classification.}
    \label{tbl:hyperparams}
    \setlength\tabcolsep{10pt}
    \centering
    \begin{tabular}{lccc}
        \toprule
        & VGG16 & ResNet-50 & Deep R-MAC \\
        \midrule
        $\gamma$ & 10.0 & 5.0 & 5.0 \\
        $C$ & 2.5 & 0.5 & 0.005 \\
        \bottomrule
    \end{tabular}
\end{table}

\section{Experiments}
\label{sec:experiments}

We evaluate the performance of the different approaches and feature representations in two scenarios:
First, we investigate a ranking task, where all images in a given dataset should be ranked according to their relevance, from the most to the least relevant.
This corresponds to a post-hoc analysis of the event, where all potentially useful images have already been collected from the web and an analyst will go through the ordered list.
This approach avoids hard decisions, which are prone to erroneously excluding relevant images, but instead allows the analysts to decide when to stop, while they go through the ranked list.

Second, we turn our attention to an on-line filtering task, which is more suitable for a \emph{rapid} flood impact analysis scenario.
In this setting, the event is still ongoing, and we need to decide for every newly posted image immediately whether it is worth looking at or irrelevant.

\subsection{Ranking Images by Relevance}
\label{subsec:exp-ranking}

The ranking scenario reflects a classical retrieval task, and we hence use established retrieval metrics for our experimental evaluation: \textit{average precision (AP)} and its mean (mAP) over the three search objectives.
AP assesses the quality of the entire ranking of all images by computing the area under the \textit{precision-recall curve}.
\textit{Recall} is the fraction of all relevant images found up to a certain position in the ranking and \textit{precision} is the fraction of images among those found up to this position that are relevant.

The results for both the European Flood 2013 dataset and our two novel Twitter datasets are given in \cref{tbl:performance}.
In addition, the precision-recall curves in \cref{fig:recprec-euflood-twitter} provide detailed insights into the performance for different recall-precision trade-offs on the European Flood 2013 test set and a combination of the two Twitter datasets.

\begin{figure}[t]
    \captionof{table}{AP for all tested methods on all datasets. Numbers are in \%. The best value per column is set in bold. Cls.~=~Classification, Ret.~=~Retrieval.}
    \label{tbl:performance}
    \tiny
    \setlength{\tabcolsep}{3pt}
    \setlength{\aboverulesep}{0pt}
    \setlength{\belowrulesep}{0pt}
    \setlength{\extrarowheight}{.75ex}
    \begin{tabularx}{\linewidth}{X cccs cccs cccs}
        \toprule
        & \multicolumn{4}{c}{European Flood 2013} & \multicolumn{4}{c}{Harz17} & \multicolumn{4}{c}{Rhine18} \\
        \cmidrule(r{3pt}){2-5}\cmidrule(l{3pt}r{3pt}){6-9}\cmidrule(l{3pt}){10-13}
        \rowcolor{white}
        \legendentry{Method}{white}
        & Flood & Depth & Poll. & mAP
        & Flood & Depth & Poll. & mAP
        & Flood & Depth & Poll. & mAP \\
        \midrule
        \legendentry{Text-Time-based}{white} & --- & --- & --- & --- & 45.6 & 30.5 & 0.6 & 25.6 & 41.7 & 21.4 & 0.3 & 21.1 \\
        \arrayrulecolor{lightgray}\midrule\arrayrulecolor{black}
        \legendentry{Cls. (VGG16)}{pltclr1} & 80.7 & 65.2 & 56.0 & 67.3 & 84.1 & 66.8 & 3.4 & 51.4 & 82.5 & \textbf{66.4} & \textbf{2.4} & \textbf{50.4} \\
        \legendentry{Cls. (ResNet-50)}{pltclr2} & \textbf{92.1} & \textbf{77.8} & \textbf{90.4} & \textbf{86.8} & 86.4 & \textbf{71.1} & \textbf{9.6} & \textbf{55.7} & \textbf{83.1} & 65.2 & 1.0 & 49.8 \\
        \legendentry{Cls. (Deep R-MAC)}{pltclr3} & 92.0 & 77.1 & 70.9 & 80.0 & \textbf{86.9} & 70.8 & 3.1 & 53.6 & 81.5 & 59.5 & 1.0 & 47.3 \\
        \arrayrulecolor{lightgray}\midrule\arrayrulecolor{black}
        \legendentry{Ret. (VGG16)}{pltclr4} & 62.0 & 53.2 & 14.0 & 43.1 & 71.2 & 58.4 & 0.5 & 43.4 & 76.2 & 63.4 & 0.7 & 46.7 \\
        \legendentry{Ret. (ResNet-50)}{pltclr5} & 69.6 & 54.2 & 37.5 & 53.8 & 64.7 & 53.7 & 1.0 & 39.8 & 75.2 & 61.0 & 1.1 & 45.8 \\
        \legendentry{Ret. (Deep R-MAC)}{pltclr6} & 62.8 & 46.4 & 28.4 & 45.9 & 83.3 & 64.6 & 2.9 & 50.3 & 80.1 & 60.0 & 1.3 & 47.1 \\
        \bottomrule
    \end{tabularx}

    \vspace*{\floatsep}

    \resizebox{\linewidth}{!}{\input{figures/recprec_euflood_twitter_nolegend.pgf}}%
    \captionof{figure}{Precision-recall curves for the European Flood 2013 test set (top row) and the combined Twitter datasets (bottom row).}
    \label{fig:recprec-euflood-twitter}
\end{figure}

On the European Flood 2013 dataset, the classification-based approach outperforms retrieval by a large margin.
ResNet-50 features work best for both approaches across all tasks.
This is an interesting finding because Barz et al.~\cite{barz2019enhancing} identified Deep R-MAC as the best representation on this dataset, but only compared it with VGG16 and not with ResNet-50.

These findings transfer well to the Harz17 dataset, where classification with ResNet-50 features performs best as well on average.
It also performs quite well on the Rhine18 dataset but is slightly outperformed there by VGG16 features, which provide surprisingly good results on that dataset only.

On both Twitter datasets, our approach efficiently filters images relevant for the flooding and depth task, but performs much worse for the pollution task than on the European Flood 2013 data.
First, there are very few relevant pollution images on Twitter (0.5\% in our dataset).
Secondly, we observed that reflections on the water surface were often mistaken for oil films.
This difficult category certainly requires more work in the future, parts of which should focus on collecting dedicated pollution image data for a more robust evaluation.

The discrepancy between the performance of the retrieval approach and the classification-based approach is not as big on the Twitter data as on the European Flood 2013 dataset, because the classifiers were trained on the domain of the latter.
That ranking by classification scores outperforms the classical retrieval approach is simply due to the closed-world scenario we employed in this work:
Retrieval as done by Barz et al.~\cite{barz2019enhancing} is more suitable for an open world, where the categories searched for by the user are not known in advance.
Restricting the filter to the three categories defined in \cref{subsec:eu-flood-dataset}, however, gives classification an advantage by being optimized for these particular tasks.

The qualitative examples shown in \cref{fig:examples} illustrate that our approach effectively filters out irrelevant images such as memes and still images from TV shows.
For each search objective, the images more relevant for that objective are ranked higher than other images of the flood.
These examples from the Harz17 dataset were generated using the classification approach with ResNet-50 features.

\subsection{On-Line Filter with Hard Decisions}
\label{subsec:exp-filter}

In the scenario of filtering a stream of incoming images, a hard decision about the relevance of individual images must be enforced.
This can be done by thresholding the scores predicted by the SVMs in case of the classification-based approach or thresholding the distance to the query images in case of retrieval.
Due to this hard decision, we only obtain one pair of recall and precision values on each of our Twitter datasets, in contrast to the retrieval setting.
We combine these by computing the so-called F1-score, which is the harmonic mean of recall and precision.
The maximum F1-scores over all possible thresholds are shown in \cref{tbl:fscore}.
This means, we assume that an optimal threshold has already been found, which usually needs to be done using held-out training data or cross-validation.

\begin{table}[t]
    \caption{Best possible F1-score (in \%) that can be obtained with each method.}
    \label{tbl:fscore}
    \small
    \setlength{\aboverulesep}{0pt}
    \setlength{\belowrulesep}{0pt}
    \setlength{\extrarowheight}{.5ex}
    \begin{tabularx}{\linewidth}{X cccs cccs}
        \toprule
        & \multicolumn{4}{c}{Harz17} & \multicolumn{4}{c}{Rhine18} \\
        \cmidrule(r{3pt}){2-5}\cmidrule(l{3pt}){6-9}
        \rowcolor{white}
        \legendentry{Method}{white}
        & Flood & Depth & Poll. & Avg
        & Flood & Depth & Poll. & Avg \\
        \midrule
        \legendentry{Text-Time-based}{white} & 59.5 & 43.7 & 3.4 & 35.5 & 59.9 & 38.6 & 0.9 & 33.1 \\
        \arrayrulecolor{lightgray}\midrule\arrayrulecolor{black}
        \legendentry{Classification (VGG16)}{pltclr1} & 78.6 & 64.6 & 20.0 & 54.4 & 78.8 & \textbf{62.4} & \textbf{16.7} & \textbf{52.6} \\
        \legendentry{Classification (ResNet-50)}{pltclr2} & 81.1 & \textbf{68.7} & \textbf{33.3} & \textbf{61.0} & \textbf{79.1} & 61.8 & 8.7 & 49.9 \\
        \legendentry{Classification (Deep R-MAC)}{pltclr3} & \textbf{81.2} & 66.5 & 12.5 & 53.4 & 76.5 & 61.1 & 5.0 & 47.5 \\
        \arrayrulecolor{lightgray}\midrule\arrayrulecolor{black}
        \legendentry{Retrieval (VGG16)}{pltclr4} & 65.0 & 55.1 & 1.6 & 40.6 & 70.6 & 60.0 & 4.1 & 44.9 \\
        \legendentry{Retrieval (ResNet-50)}{pltclr5} & 58.4 & 52.0 & 5.4 & 38.6 & 69.5 & 58.1 & 7.1 & 44.9 \\
        \legendentry{Retrieval (Deep R-MAC)}{pltclr6} & 77.3 & 64.9 & 9.3 & 50.5 & 76.3 & 57.8 & 5.6 & 46.6 \\
        \bottomrule
    \end{tabularx}
\end{table}

In this scenario, the superiority of the classification-based approach is more pronounced as in the retrieval setting.
Again, we can observe that VGG16 features only perform well on the Rhine18 data, while ResNet-50 features provide best or competitive performance for both datasets.
Averaged over both datasets, classification with ResNet-50 features achieves a precision of 73\% and a 89\% recall for the flooding task and a precision of 65\% with a recall of 67\% for the depth task.
This illustrates the benefit of using machine learning for filtering relevant social media images, since only every fourth image passing the filter will not show flooding, while still 89\% of all relevant images are found.
If one would aim for 99\% recall, the precision on the flooding task would still be 54\%.

\section{Conclusions}
\label{sec:conclusions}

We presented an automatic filter for images posted on Twitter with respect to their relevance for obtaining various information about floodings and rapidly assessing flood impacts, so that response and recovery can be coordinated quickly and adequately.
To this end, we have shown that classifiers trained on data from Wikimedia Commons can be applied successfully to real Twitter data.
While retrieval-based approaches used in the past are flexible and enable the user to refine the results easily by giving feedback, classification is faster, does not require interactivity, and provides better filtering performance, which makes it more suitable for gaining insights rapidly during the event from streaming data.
Thus, we recommend our classification model based on ResNet-50 features for use in practice, since it provides best or at least competitive performance across tasks and datasets.
A realistic application scenario, however, poses further challenges:
Since many images posted on Twitter lack accurate geodata, a technique for automatic geolocalization of tweets or images is crucial.
In this work, we focused on finding relevant images, and leave the geolocalization to future work.

\subsubsection*{Acknowledgements}

This work was supported by the German Research Foundation as part of the
programme ``Volunteered Geographic Information: Interpretation,
Visualisation and Social Computing'' (SPP 1894, contract DE 735/11-1).

\bibliographystyle{splncs04}
\bibliography{references}

\end{document}